\pdfoutput=1
\documentclass[11pt]{article}

\usepackage{styles/acl}

\usepackage{times}
\usepackage{latexsym}

\usepackage[T1]{fontenc}
\usepackage[utf8]{inputenc}

\usepackage{microtype}

\usepackage{inconsolata}

\usepackage{graphicx} 
\graphicspath{ {./images/} }
\usepackage{algorithm} 
\usepackage{algpseudocode}

\title{MrRank: Improving Question Answering Retrieval System through Multi-Result Ranking Model}

\author{
Danupat Khamnuansin\textsuperscript{\textdagger*},
Tawunrat Chalothorn\textsuperscript{\textdagger}, \textnormal{and} Ekapol Chuangsuwanich\textsuperscript{*} \\ 
\textsuperscript{\textdagger} Kasikorn Labs Co., Ltd., Kasikorn Business-Technology Group, Thailand \\
\textsuperscript{*} Department of Computer Engineering, Faculty of Engineering, Chulalongkorn University \\
\small{\texttt{danupat.k@kbtg.tech \quad{tawunrat.c@kbtg.tech} \quad{ekapolc@cp.eng.chula.ac.th}}}}

\begin{document}
\maketitle
\begin{abstract}
Large Language Models (LLMs) often struggle with hallucinations and outdated information. To address this, Information Retrieval (IR) systems can be employed to augment LLMs with up-to-date knowledge. However, existing IR techniques contain deficiencies, posing a performance bottleneck. Given the extensive array of IR systems, combining diverse approaches presents a viable strategy. Nevertheless, prior attempts have yielded restricted efficacy. In this work, we propose an approach that leverages learning-to-rank techniques to combine heterogeneous IR systems. We demonstrate the method on two Retrieval Question Answering (ReQA) tasks. Our empirical findings exhibit a significant performance enhancement, outperforming previous approaches and achieving state-of-the-art results on ReQA SQuAD.

\end{abstract}

\section{Introduction}

Large Language Models (LLMs) have demonstrated proficiency in comprehending human language across diverse domains. Among these advancements, question answering (QA) systems integrated with LLMs have gained significant attention. Nonetheless, concerns persist regarding the susceptibility of LLMs to hallucinations, potentially leading to inaccuracies in their responses.

To address the aforementioned challenge, the Retrieval Augmented Generation (RAG) \cite{10.5555/3495724.3496517} is proposed. The RAG system generally consists of two parts: an Information Retrieval (IR) engine which proposes the top $k$ candidate pieces from a document corpus for the query, and the QA engine which is responsible for deriving the final answer from retrieved information.

Although LLMs demonstrate exceptional proficiency in QA tasks, contemporary text embedding techniques employed in IR remain underdeveloped. This deficiency poses a performance bottleneck for RAG, as shown by the results in the Massive Text Embedding Benchmark \cite{muennighoff-etal-2023-mteb}.

To improve retrieval performance, advanced fine-tuning techniques applied to individual retrieval systems \cite{10082873} or the enhancement of model components \cite{zhao-etal-2021-sparta} emerge as a prominent area of investigation. While fine-tuning approaches have demonstrated efficacy in this domain, they commonly require substantial resources for training, posing a challenge to real-world adoption.

Combining results from multiple IR systems can be quite effective, as demonstrated in the work of \citet{10.1145/1571941.1572114}, Reciprocal Rank Fusion (RRF) is a technique that utilizes ranking information to rerank the overall ranking, which has become the industry standard for combining multiple ranking systems. Although the implementation of RRF is straightforward, a notable limitation arises due to its failure to leverage output scores from individual retrievers. Modern techniques have been developed to overcome this limitation; FlexNeuART \cite{boytsov-nyberg-2020-flexible} employs the Learning to Rank (LTR) framework to amalgamate scores from diverse retrievers into a unified score. However, the application of LTR in the QA retrieval setting requires a modification of the training process to avoid irrelevant samples. Moreover, the exploration and optimization of these techniques within the domain of question answering retrieval settings remain largely unaddressed. 

In the work of \citet{10.1007/978-3-030-72113-8_22}, a routing method for combining multiple retrievers in the QA setting is introduced. This model employs a routing classifier to select between a neural retriever and a traditional BM25 retriever. However, this method heavily relies on the score generated by BM25 and disregards ranking information from the neural model. Consequently, its performance is limited in modern question answering systems where the neural model typically outperforms BM25.

We propose re-ranking techniques specifically trained to improve question answering retrieval performance. Our investigations involve combining two different types of model architecture (term weighting and neural networks). Additionally, we conducted experiments on two distinct styles of ReQA datasets to demonstrate that combining multiple models using this setup can leverage the advantages of both models through ranking loss objective training. Our findings show that the proposed methodology yields enhanced retrieval performance and ultimately improves the downstream QA task. Our approach outperforms the current state-of-the-art \cite{10082873} on ReQA SQuAD, surpassing all individual retrieval models, RRF, and the statistical routing strategy, yielding an average enhancement of 13.6\% in the mean reciprocal rank (MRR) across datasets.

\section{Related Work}

\subsection*{Massive Text Embedding Benchmark}
Massive Text Embedding Benchmark (MTEB) proposed by \citet{muennighoff-etal-2023-mteb} offers comprehensive insights and evaluations across a wide range of modern text embedding models. This benchmark is particularly crucial for understanding the out-of-the-box performance of various models. The evaluation incorporated established text embedding models available on the Hugging Face platform \cite{wolf-etal-2020-transformers}. The assessment encompassed retrieval tasks across 15 distinct datasets, each representing diverse domains and dataset characteristics. The evaluation of re-ranking tasks was carried out using four specifically curated re-ranking datasets, providing a robust assessment of performance in real-world applications.

As shown in \autoref{tab:table_1}, GPT Sentence Embeddings for Semantic Search (SGPT) \citep{muennighoff2022sgpt} is one of the top performing models in retrieval tasks. Note that sentence embeddings constructed by simply averaging Global Vectors for Word Representation (GloVe) \cite{pennington-etal-2014-glove} outperforms BERT \cite{devlin-etal-2019-bert} and OpenAI's Ada Similarity. This observation underscores the challenges inherent in retrieval tasks, where mainstream text embedding models still exhibit a deficiency in text retrieval proficiency.

In the context of re-ranking tasks, MPNET \citep{10.5555/3495724.3497138} has demonstrated superior performance, highlighting the challenges in achieving a universally best solution in the current landscape. Notably, a model exhibiting proficiency in retrieval tasks may not necessarily excel in re-ranking tasks. 

\begin{table}[H]
\centering
\begin{tabular}{l c c c c}
\hline
\textbf{Model} & \textbf{Retr.} & \textbf{Rerank.} \\
\hline
{BERT} & {10.59} & {43.44} \\
{OpenAI Ada Similarity} & {18.36} & {49.02} \\
{GloVe} & {21.62} & {43.29} \\
{MPNet} & \underline{43.81} & \textbf{59.36} \\
{SGPT-5.8B-msmarco} & \textbf{50.25} & \underline{56.56} \\
\hline
\end{tabular}
\caption{Average of the retrieval and re-ranking metric per task per model on MTEB English subsets.}
\label{tab:table_1}
\end{table}

\subsection*{Retrieval Question Answering}
Retrieval question answering (ReQA) tasks are proposed in works such as \citet{ahmad-etal-2019-reqa} and \citet{chen-etal-2017-reading}. They encompass a type of information retrieval system that takes a question as input and responds with the most relevant sentence containing the answer. This task distinguishes itself from traditional Question Answering (QA), where the paragraph containing the answer is provided upfront, leading to notable achievements in machine learning systems, some of which surpass human performance \citep{rajpurkar-etal-2016-squad}. QA retrieval systems typically rely on either neural or term weighting methods. Recent studies, such as SPARTA \citep{zhao-etal-2021-sparta}, exhibit significant improvements in retrieval performance by adhering to the concept of tokens-level matching from traditional IR systems. However, the exploration of QA retrieval systems that incorporate hybrid retrieval mechanisms remains relatively limited.

\subsection*{Hybrid Method for Evidence Retrieval for Question Answering}

Previous attempts to combine a term weighting model with a neural-based model simply involve routing between retrieval results from each model. This is achieved by training a routing classifier that determines whether to use the traditional IR method (BM25) or the neural-based method (USE-QA) based on statistics derived from the training dataset \cite{10.1007/978-3-030-72113-8_22}. The study demonstrates that the hybrid approach outperforms individual retrieval strategies in ReQA SQuAD and ReQA NQ. However, this methodology has limitations in its support for combining multiple neural models, the integration of more than two models, and heavily relies on BM25. These limitations may deter real-world adoption, where there are numerous retrieval models specialized in their own areas. Notably, USE-QA \cite{yang-etal-2020-multilingual}, employs fine-tuning techniques to tackle an asymmetric matching task present in QA retrieval. Combining the specialized abilities of these models could lead to better performance.

Thus, we emphasized an approach enhancing model retrieval performance through model-level combination that aims to capitalize on the strengths of multiple model types. Our objective is to present a robust methodology that can effectively combine state-of-the-art or fine-tuned models for better performance improvement.

\begin{figure}[ht]
\centering
\caption{Our framework for amalgamating multiple ranking information by using the re-ranking approach.}
\includegraphics[width=7.25cm]{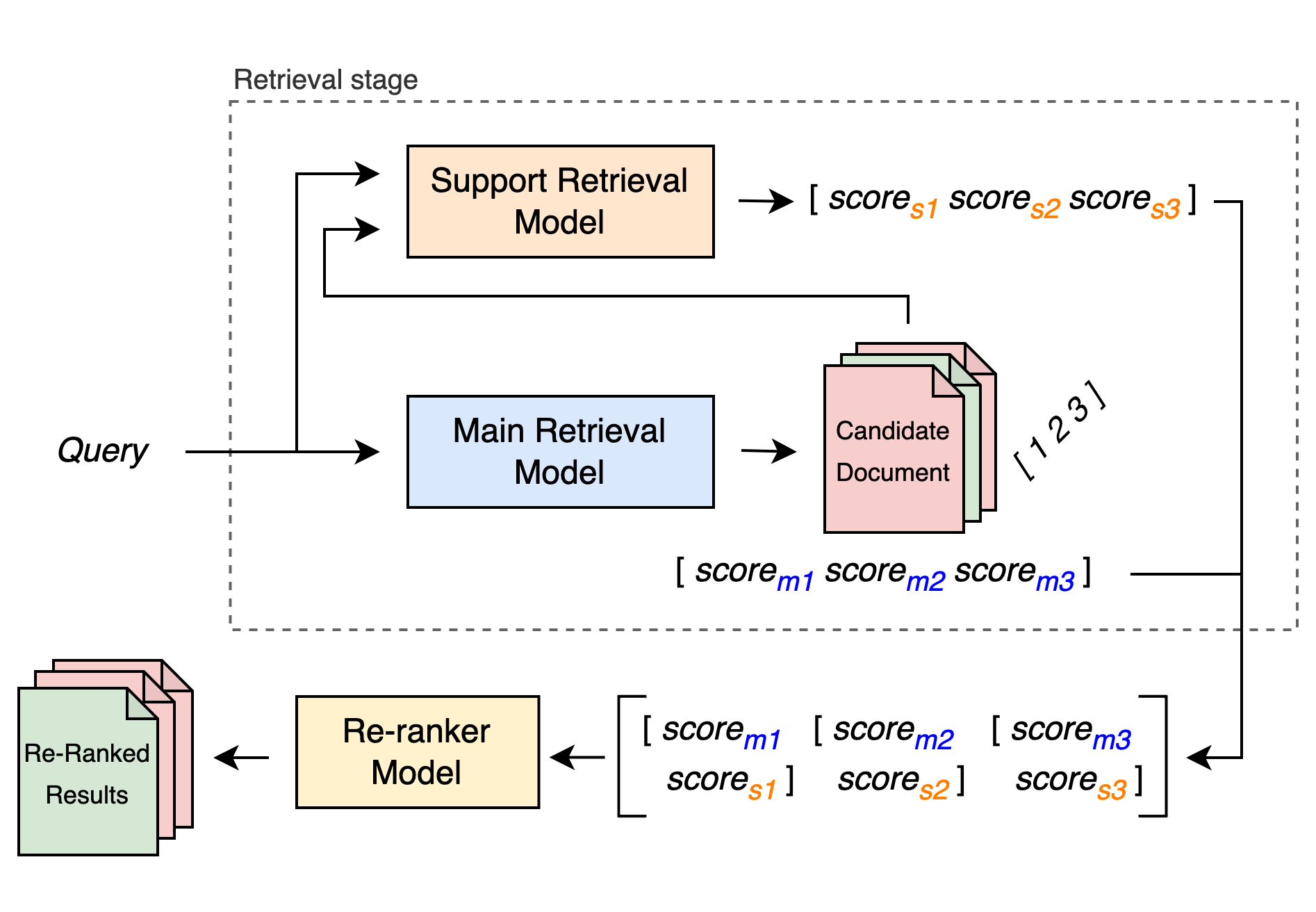}
\label{fig:figure_1}
\end{figure}

\section{Proposed Method}

To construct a cohesive framework for merging multiple sets of ranking information, we adopt a re-ranking approach. We have designated a neural-based model as the primary retriever, complemented by BM25 as a supporting model. This choice is based on the observed consistency of the neural-based model, which demonstrates superior performance on R@10. This metric is particularly crucial for assessing the upper limits of performance gain in re-ranking performance. The rankings obtained from the main retriever serve as the initial rankings, subsequently augmented by incorporating scores derived from the support retriever. These concatenated scores are then fed into the re-ranker model to predict the final rankings. The overall process is illustrated in \autoref{fig:figure_1}.

The frameworks consist of two stages:

1) Retrieval Stage: off-the-shelf retrievers are used in order to generate a candidate pool. We select top candidate retrievers from MTEB benchmarks within the retrieval and re-ranking category. The performance is shown in \autoref{tab:table_2}.

2) Re-ranking Stage: a re-ranking network is used to construct the final ranking from the candidate pool.

\subsection{Retrieval Stage}
Our approach focuses on incorporating multiple QA retrieval models. These models have demonstrated significant performance improvements in question answering retrieval tasks compared to base retrieval models. To maximize the applicability of our approach, we have chosen widely recognized and publicly accessible models. In accordance with \citet{muennighoff-etal-2023-mteb} results, we also included SGPT-5.8B-msmarco and MPNET. These models represent the current state-of-the-art publicly accessible models in the domains of retrieval and re-ranking tasks, respectively.

\begin{table}[ht]
\centering
\begin{tabular}{l c c c c}
\hline
\textbf{Model} & \textbf{MRR} & \textbf{R@1} & \textbf{R@5} & \textbf{R@10}\\
\hline
\multicolumn{5}{l}{\it ReQA SQuAD} \\
{BM25} & \underline{0.670} & \underline{0.591} & {0.759} & {0.814} \\
{USE\textsubscript{qa}} & {0.665} & {0.561} & \underline{0.793} & \underline{0.854}  \\
{MPNET\textsubscript{qa}} & {0.549} & {0.399} & {0.748} & {0.847}  \\
{SGPT\textsubscript{5.8B}} & \textbf{0.783} & \textbf{0.699} & \textbf{0.887} & \textbf{0.926}  \\
\hline
\multicolumn{5}{l}{\it ReQA NQ} \\
{BM25} & {0.529} & {0.378} & {0.723} & {0.797} \\
{USE\textsubscript{qa}} & {0.582} & \underline{0.447} & {0.751} &  {0.826} \\
{MPNET\textsubscript{qa}} & \underline{0.628} & {0.439} & \textbf{0.902} & \textbf{0.950} \\
{SGPT\textsubscript{5.8B}} & \textbf{0.652} & \textbf{0.528} & \underline{0.807} & \underline{0.856}  \\
\hline
\end{tabular}
\caption{The off-the-shelf sentence-level retrieval performance on the ReQA SQuAD and ReQA NQ task.}
\label{tab:table_2}
\end{table}

\textbf{BM25 Retriever}
The token statistics for BM25 in the original ReQA paper were computed over the first 10,000 questions of each dataset. This led to significant performance losses, as shown in \autoref{tab:table_3}. In our implementation of BM25, we have made modifications that include the following key distinctions:

1) The statistics are derived from all documents within the training set, instead of solely utilizing the first 10,000 questions.

2) Our BM25 model is constructed using the gensim library \citep{rehurek2011gensim} with default parameters.

The original ReQA paper provided the rationale behind using only the initial 10,000 questions as to reduce the scoring time. We found that there is no discernable disparity in inference time when comparing the utilization of all question statistics versus using only the first 10,000 statistics for constructing BM25. Consequently, we recommend adopting our approach to implement the BM25 version instead, which yields a notable increase in performance.
It is worth mentioning that the training and inference duration of our model remains unaffected by these variations in BM25 configurations.

\begin{table}
\centering
\begin{tabular}{r c c c c}
\hline
\textbf{Model} & \textbf{MRR} & \textbf{R@1} & \textbf{R@5} & \textbf{R@10}\\
\hline
\multicolumn{5}{l}{\it ReQA SQuAD} \\
{BM25\textsubscript{10k}} & {0.600} & {0.523} & {0.683} & {0.741}  \\
{BM25\textsubscript{all}} & \textbf{0.670} &  \textbf{0.591} &  \textbf{0.759} & \textbf{0.814} \\
\hline
\multicolumn{5}{l}{\it ReQA NQ} \\
{BM25\textsubscript{10k}} & {0.474} & {0.327} & {0.660} & {0.739}  \\
{BM25\textsubscript{all}} & \textbf{0.529} &  \textbf{0.378} &  \textbf{0.723} & \textbf{0.797} \\
\hline
\end{tabular}
\caption{Performances of BM25 on sentence-level ReQA from original paper and our implementation.}
\label{tab:table_3}
\end{table}

\subsection{Re-ranking Stage}
To effectively integrate ranking information derived from multiple retrieval systems, we opted for a neural-based architecture with a pairwise learning-to-rank approach. The method is motivated by two primary factors.

1) QA retrieval models are initially pre-trained with a specific focus on enhancing the ranking of question-answer document pairs. Consequently, applying the pointwise method to train the re-ranker model would lead to redundant training.

2) The unavailability of comprehensive ranking in the question-answer retrieval dataset poses limitations on our ability to fully utilize the advantageous attributes of listwise methods.

\textbf{Architecture}
The re-ranker model adopts the RankNet \citep{10.1145/1102351.1102363} architecture, which comprises of two layers of Siamese feed-forward neural network. \citep{koch2015siamese} The network consists of 10 hidden units with leaky ReLU \citep{maas2013rectifier} and a single sigmoid output node.

\textbf{Training Hyperparameters}
We optimized the re-ranker network using the Adam optimizer, \citep{DBLP:journals/corr/KingmaB14} using a learning rate of $1e-3$, 1024 batch size, and 100 epochs for every model setting.

\textbf{Data Preparation}
The training features are derived from the top 64 ranking scores of the best-performing retrieval model during the retrieval stage (main retriever). Subsequently, the scores from support models are computed and concatenated to form a set of re-ranking features.
For the term weighting model, we utilize the scoring output directly from the model. For the neural-based model, the score is generated by calculating the cosine similarity between the encoded query and the candidate document vectors.

We chose a number of re-ranked documents ($k$) equal to 64 to facilitate a direct comparison with the routing method. We experimented on how the value of $k$ affects the performance and inference time. The findings indicate that augmenting the value of $k$ beyond 16 yields only marginal improvements in MRR and is not worth trading off with prediction speed. Detailed experimental outcomes are presented in \autoref{fig:figure_4} and \autoref{fig:figure_5}, as outlined in the Appendix.

\begin{figure*}
% \centering
\caption{Data preparation steps include: 1) retrieving scores for each document candidate, 2) performing permutations on each pair of documents and filtering out tied pairs, and 3) creating negative and positive samples.}
\includegraphics[width=15.5cm]{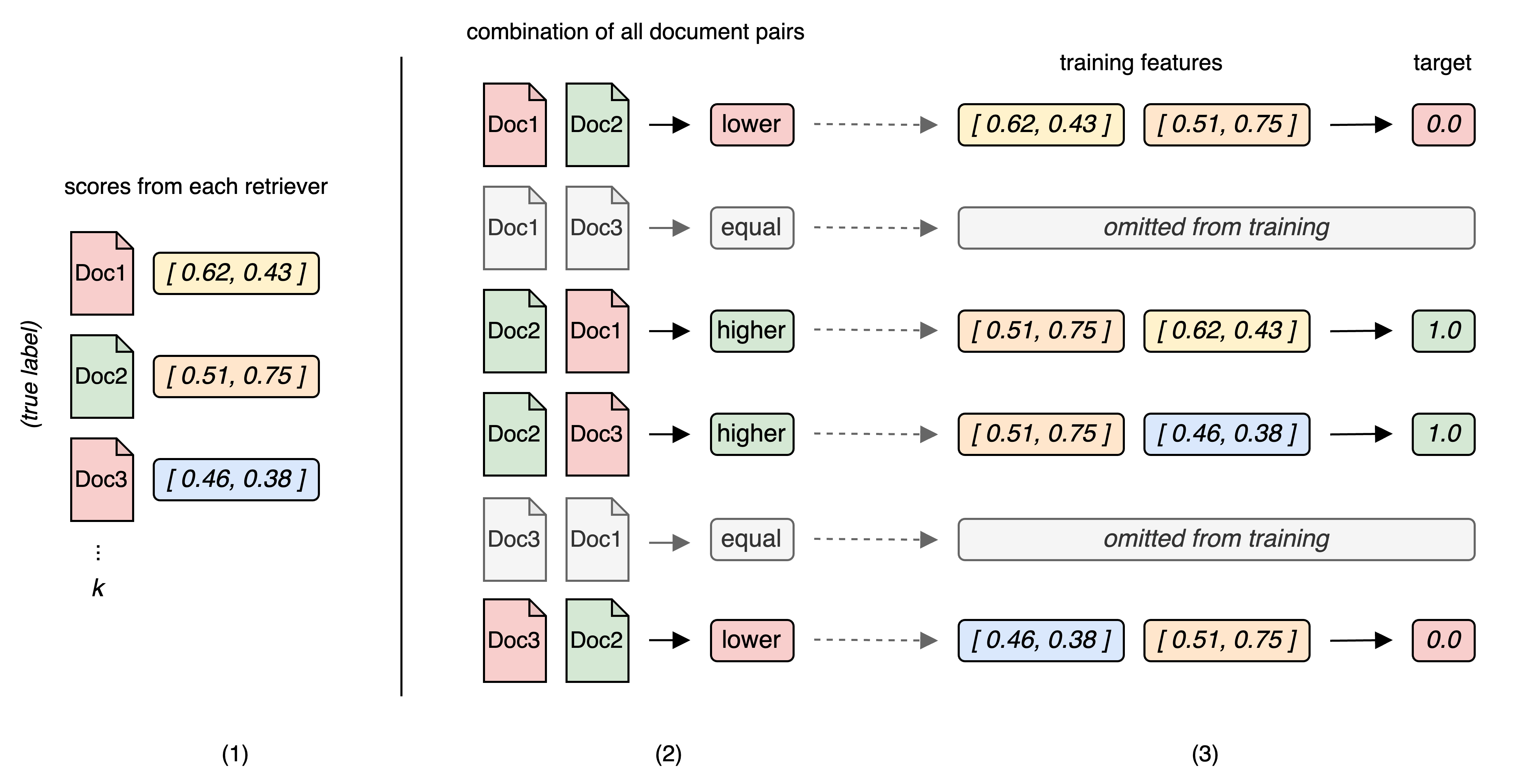}
\label{fig:figure_3}
\end{figure*}

\textbf{Training Objective}
The pairwise LTR framework objective involves training the model to produce a value closer to one when it determines the ranking of a particular document to be higher than another. Conversely, it generates a value closer to zero for the opposite scenario. In the event of a tied situation, the conventional RankNet framework designates a target value of 0.5. Since the ReQA dataset has a lot of non-target documents, selecting random pairs will usually result in a tie. To customize the training objective for ReQA tasks, our approach prioritizes sentences containing the correct answer over those that do not. We have modified the conventional RankNet learning-to-rank algorithm by excluding the training sample with a target value of 0.5. The reduction in unnecessary training pairs resulted in a significant decrease in training time while simultaneously enhancing performance.

\textbf{Inference Phase} During the inference phase, the re-ranker model generates ranking information for each candidate document pair to determine which document should rank higher \{(D\textsubscript{i} > D\textsubscript{j}), (D\textsubscript{j} > D\textsubscript{k}), (D\textsubscript{i} > D\textsubscript{k})\}. The final ranking is derived using a topological sorting algorithm \{D\textsubscript{i} > D\textsubscript{j} > D\textsubscript{k}\}. In cases where the outcome fails to yield a conclusive ranking, the original results from the main retriever are used as the final ranking. All operations are performed using the networkx library \citep{SciPyProceedings_11}.

\section{Results}

We assess different retrieval models, including BM25, USE-QA, SGPT-5.8B-msmarco and MPNET-QA \citep{10.5555/3495724.3497138}. Our baselines include average scoring, thresholding, and Rank Reciprocal Fusion (RRF), a technique commonly employed within modern multi-retriever systems.

We compare our proposed approach mainly on the ReQA SQuAD and ReQA NQ, aligning with the original ReQA paper. Additionally, we conduct experiments on seven question answering datasets from MTEB to assess its effectiveness in contemporary retrieval tasks.

The Stanford Question Answering Dataset (SQuAD) \cite{rajpurkar-etal-2016-squad} consists of questions on a set of Wikipedia articles, whereas Natural Questions (NQ) \cite{kwiatkowski-etal-2019-natural} involves user questions issued to Google search, with corresponding answers sourced from Wikipedia by annotators. These two datasets represent distinct characteristics related to lexical overlap between questions and contexts, with SQuAD representing a low-overlap scenario while NQ exhibits the opposite. The publicly available version of both datasets consists of two partitions: a training set and a development set. The number of questions and documents is shown in \autoref{tab:table_4}.

The development set was used as the evaluation set for both datasets. We focused on the sentence-level retrieval task, following the method outlined in the original ReQA paper. However, our implementation of BM25 included the complete vocabulary of the training documents, deviating slightly from the original ReQA paper which only used the first 10,000 questions. This difference resulted in better performance for our BM25 implementation. See "BM25 Retriever" in Section 3.1 for additional details.

\begin{table}[H]
\centering
\begin{tabular}{l r r}
\hline
\textbf{Dataset} & \textbf{Question} & \textbf{Document} \\
\hline
{ReQA\textsubscript{SQuAD} (train)} & {87,599} & {91,707} \\
{ReQA\textsubscript{SQuAD} (dev)} & {11,426} & {10,250} \\
{ReQA\textsubscript{NQ} (train)} & {74,097} & {239,013} \\
{ReQA\textsubscript{NQ} (dev)} & {1,772} & {7,020} \\
\hline
\end{tabular}
\caption{The number of questions and candidate sentences on the ReQA SQuAD and ReQA NQ datasets.}
\label{tab:table_4}
\end{table}

\begin{table*}[t]
\centering
\begin{tabular}{r c l c c c}
\hline
\textbf{Model} & \textbf{Type} & \textbf{MRR} & \textbf{R@1} & \textbf{R@5} & \textbf{R@10}\\
\hline
\multicolumn{6}{l}{\it ReQA SQuAD} \\
{BM25} & {Single} & {0.670} & {0.591} & {0.759} & {0.814} \\
{USE\textsubscript{qa}} & {Single} & {0.665} & {0.561} & {0.793} & {0.854}  \\
{MPNET\textsubscript{qa}} & {Single} & {0.549} & {0.399} & {0.748} & {0.847}  \\
{SGPT\textsubscript{5.8B}} & {Single} & {0.783} & {0.699} & {0.887} & {0.926}  \\
{RRF\textsubscript{use} (USE\textsubscript{qa}+BM25)} & {Hybrid} & {0.721} & {0.624} & {0.839} & {0.895} \\
{Routing\textsubscript{use} (USE\textsubscript{qa}+BM25)} & {Hybrid} & {0.705} & {0.626} & {0.796} & {0.851} \\
{Ours\textsubscript{use} (USE\textsubscript{qa}+BM25)} & {Hybrid} & {0.750} & {0.663} & {0.860} & {0.902} \\
{RRF\textsubscript{mp} (MPNET\textsubscript{qa}+BM25)} & {Hybrid} & {0.673} & {0.546} & {0.827} & {0.896} \\
{Routing\textsubscript{mp} (MPNET\textsubscript{qa}+BM25)} & {Hybrid} & {0.709} & {0.623} & {0.811} & {0.869} \\
{Ours\textsubscript{mp} (MPNET\textsubscript{qa}+BM25)} & {Hybrid} & {0.748} & {0.658} & {0.856} & {0.909} \\
{RRF\textsubscript{sgpt} (SGPT\textsubscript{5.8B}+BM25)} & {Hybrid} & {0.754} & {0.664} & {0.864} & {0.914} \\
{Routing\textsubscript{sgpt} (SGPT\textsubscript{5.8B}+BM25)} & {Hybrid} & {0.762} & {0.686} & {0.852} & {0.897} \\
{Ours\textsubscript{sgpt} (SGPT\textsubscript{5.8B}+BM25)} & {Hybrid} & \textbf{0.815} & \textbf{0.736} & \textbf{0.913} & \textbf{0.945} \\
\hline
\multicolumn{6}{l}{\it ReQA NQ}\\
{BM25} & {Single} & {0.529} & {0.378} & {0.723} & {0.797} \\
{USE\textsubscript{qa}} & {Single} & {0.582} & {0.447} & {0.751} &  {0.826} \\
{MPNET\textsubscript{qa}} & {Single} & {0.628} & {0.439} & {0.902} & {0.950} \\
{SGPT\textsubscript{5.8B}} & {Single} & {0.652} & {0.528} & {0.807} & {0.856}  \\
{RRF\textsubscript{use} (USE\textsubscript{qa}+BM25)} & {Hybrid} & {0.603} & {0.450} & {0.802} & {0.874} \\
{Routing\textsubscript{use} (USE\textsubscript{qa}+BM25)} & {Hybrid} & {0.552} & {0.396} & {0.758} & {0.831} \\
{Ours\textsubscript{use} (USE\textsubscript{qa}+BM25)} & {Hybrid} & {0.590} & {0.452} & {0.765} & {0.832} \\
{RRF\textsubscript{mp} (MPNET\textsubscript{qa}+BM25)} & {Hybrid} & {0.617} & {0.449} & {0.843} & {0.897} \\
{Routing\textsubscript{mp} (MPNET\textsubscript{qa}+BM25)} & {Hybrid} & {0.638} & {0.467} & {0.873} & {0.931} \\
{Ours\textsubscript{mp} (MPNET\textsubscript{qa}+BM25)} & {Hybrid} & {0.642} & {0.457} & \textbf{0.908} & \textbf{0.951} \\
{RRF\textsubscript{sgpt} (SGPT\textsubscript{5.8B}+BM25)} & {Hybrid} & {0.601} & {0.443} & {0.791} & {0.872} \\
{Routing\textsubscript{sgpt} (SGPT\textsubscript{5.8B}+BM25)} & {Hybrid} & {0.600} & {0.440} & {0.813} & {0.876} \\
{Ours\textsubscript{sgpt} (SGPT\textsubscript{5.8B}+BM25)} & {Hybrid} & \textbf{0.655} & \textbf{0.529} & {0.816} & {0.862} \\
\hline
\end{tabular}
\caption{Model performances on the sentence-level retrieval of ReQA SQuAD and ReQA NQ task. The routing model refers to the Hybrid (BM25) method described in the original paper.}
\label{tab:table_5}
\end{table*}

We use the Mean Reciprocal Rank (MRR) and Recall at k (R@k) as evaluation metrics. The experiments were replicated five times, and the mean value is reported. We performed significant testing between routing and our method, which showed significant improvements in all settings (one sample t-test, with p < 0.05). See Appendix for summary statistics of the results.

\autoref{tab:table_5} presents the performance of both individual and combined models. Our model demonstrates better performance across all datasets and evaluation metrics, particularly in ReQA SQuAD where BM25's performance is comparable to those of the neural-based models. Our approach exhibits a significant improvement compared to individual retrieval methods, RRF and routing strategies. Our model (SGPT+BM25) outperforms MPNET-QA by 48.5\%, USE-QA by 22.6\%, BM25 by 21.6\%, SGPT-5.8B-msmarco by 4.1\%, and routing techniques on the same model combination by 7.0\%.

On the ReQA NQ dataset, the performance of BM25 is significantly inferior compared to the neural-based systems. Our approach demonstrates the capability to effectively integrate ranking information, resulting in an average performance uplift of 2.6\% from individual neural models. This highlights the efficacy of our approach, even in situations where less potent models are combined with their more superior counterparts.

\subsection*{Common Ensemble Techniques}
We assess standard ensemble techniques including

1) Average scoring, re-rank outputs based on normalized average scores from each retriever.

2) Thresholding with a predefined criterion set at 0.5, routing between main and support retriever.

Our results indicate that common ensemble techniques demonstrate inferior performance compared to our approach across all settings, as shown in \autoref{tab:table_21}.

\begin{table}[H]
\centering
\begin{tabular}{r c c c}
\hline
\textbf{Model} & \textbf{MRR} & \textbf{R@1} & \textbf{R@10}\\
\hline
\multicolumn{4}{l}{\it ReQA SQuAD} \\
{Avg (USE+BM25)} & {0.687} & {0.595} & {0.869}  \\
{Avg (MPNET+BM25)} & {0.687} & {0.599} & {0.859}  \\
{Avg (SGPT+BM25)} & {0.696} & {0.598} & {0.901}  \\
{Thr (USE+BM25)} & {0.691} & {0.594} & {0.868}  \\
{Thr (MPNET+BM25)} & {0.571} & {0.424} & {0.861}  \\
{Thr (SGPT+BM25)} & {0.788} & {0.707} & {0.927}  \\
\hline
\multicolumn{4}{l}{\it ReQA NQ} \\
{Avg (USE+BM25)} & {0.345} & {0.198} & {0.691}  \\
{Avg (MPNET+BM25)} & {0.357} & {0.209} & {0.720}  \\
{Avg (SGPT+BM25)} & {0.346} & {0.190} & {0.720}  \\
{Thr (USE+BM25)} & {0.595} & {0.455} & {0.846}  \\
{Thr (MPNET+BM25)} & {0.630} & {0.441} & {0.949}  \\
{Thr (SGPT+BM25)} & {0.650} & {0.528} & {0.853}  \\
\hline
\end{tabular}
\caption{Retrieval performances of common ensemble techniques. $Avg$ and $Thr$ denote average scoring and thresholding, respectively.}
\label{tab:table_21}
\end{table}

\subsection*{Neural Model Combinations}
Our experiments were assessed within the framework of term weighting-neural combination in direct comparison with the routing method. We also included the neural-neural combination experiments. The results demonstrate the efficacy of our approach. The MPNET-QA and SGPT-5.8B-msmarco combination demonstrated notable improvements, surpassing the best performer in the main experiments on ReQA NQ. This is consistent with our prior observations that the fusion of competitive models yields greater performance gains. The results of the model combination are shown in \autoref{tab:table_7}.

\begin{table}[H]
\centering
\begin{tabular}{r c c c}
\hline
\textbf{Model} & \textbf{MRR} & \textbf{R@1} & \textbf{R@10}\\
\hline
\multicolumn{4}{l}{\it ReQA SQuAD} \\
{USE\textsubscript{qa} + SGPT} & {0.784} & {0.704} & {0.920}  \\
{MPNET\textsubscript{qa} + SGPT} & {0.786} & {0.696} & {0.939}  \\
{SGPT + USE\textsubscript{qa}} & {0.793} & \textbf{0.710} & {0.934}  \\
{MPNET\textsubscript{qa} + USE\textsubscript{qa}} & {0.741} & {0.634} & {0.924}  \\
{USE\textsubscript{qa} + MPNET\textsubscript{qa}} & {0.742} & {0.639} & {0.915}  \\
{SGPT + MPNET\textsubscript{qa}} & \textbf{0.799} & \textbf{0.710} & \textbf{0.948}  \\
\hline
\multicolumn{4}{l}{\it ReQA NQ} \\
{USE\textsubscript{qa} + SGPT} & {0.660} & {0.536} & {0.867}  \\
{MPNET\textsubscript{qa} + SGPT} & \textbf{0.723} & {0.569} & \textbf{0.963}  \\
{SGPT + USE\textsubscript{qa}} & {0.665} & {0.538} & {0.876}  \\
{MPNET\textsubscript{qa} + USE\textsubscript{qa}} & {0.712} & {0.551}  & {0.961}  \\
{USE\textsubscript{qa} + MPNET\textsubscript{qa}} & {0.698} & {0.553} & {0.920}  \\
{SGPT + MPNET\textsubscript{qa}} & {0.714} & \textbf{0.576} & {0.922}  \\
\hline
\end{tabular}
\caption{The performances of neural-neural models.}
\label{tab:table_7}
\end{table}

\subsection*{Overall Question Answering Performance}
In this experiment we investigated how enhancing retrieval performance translates to the overall question answering system performance. We integrated retrieval systems with the BERT reader model and found that an increase in retrieval performance benefits the overall QA system performance in every setting. We reported answer quality using two metrics, including the Exact Match score (EM), where the answer is strictly required to match the label, along with the less strict ROUGE-L score \cite{lin-2004-rouge}. The results show a strong correlation coefficient of 0.865 between R@1 and ROUGE-L scores, confirming that enhancements in retrieval performance positively impact overall QA performance. However, there is a substantial gap between the best retrieval method and the oracle retriever, showing the potential gain for further improvements. The results are presented in \autoref{tab:table_20}.

\begin{table}[H]
\centering
\begin{tabular}{r c c}
\hline
\textbf{Model} & \textbf{ROUGE-L} & \textbf{EM}\\
\hline
\multicolumn{3}{l}{\it SQuAD} \\
{BM25} & {0.519} & {0.392}  \\
{MPNET} & {0.370} & {0.264}  \\
{SGPT} & {0.611} & {0.458}  \\
{Ours (MPNET+BM25)} & {0.573} & {0.431}  \\
{Ours (SGPT+BM25)} & {0.636} & {0.481}  \\
{Ours (SGPT+MPNET)} & {0.618} & {0.466}  \\
{Ours (Trio*)} & \textbf{0.641} & \textbf{0.485}  \\
{Oracle Retriever} & {0.848} & {0.651}  \\
\hline
\multicolumn{3}{l}{\it NQ} \\
{BM25} & {0.340} & {0.235}  \\
{MPNET} & {0.392} & {0.259}  \\
{SGPT} & {0.450} & {0.311}  \\
{Ours (MPNET+BM25)} & {0.403} & {0.271}  \\
{Ours (SGPT+BM25)} & {0.450} & {0.310}  \\
{Ours (SGPT+MPNET)} & \textbf{0.488} & {0.339}  \\
{Ours (Trio*)} & \textbf{0.488} & \textbf{0.340}  \\
{Oracle Retriever} & {0.781} & {0.588}  \\
\hline
\end{tabular}
\caption{Overall performances on SQuAD and NQ. *Trio model comprises SGPT, MPNET, and BM25.}
\label{tab:table_20}
\end{table}

\subsection*{Effect of Sample Mining Technique}
We trained the the combination model of BM25 and SGPT-5.8B-msmarco using the original pairwise learning-to-rank approach, and compared it to our modified version which includes sample mining. \autoref{tab:table_6} shows that by reducing unnecessary training pairs, there is a substantial 97.0\% reduction in training samples. At the same time, it yields an average 0.6\% marginally improvement in performance across datasets.

\begin{table}[H]
\centering
\begin{tabular}{r c c}
\hline
\textbf{Model} & \textbf{MRR} & \textbf{Training Pair}\\
\hline
\multicolumn{3}{l}{\it ReQA SQuAD} \\
{Ours\textsubscript{sgpt}} & \textbf{0.8146 \textpm0.0004} & \textbf{5,212,691} \\
{-mining\textsubscript{sgpt}} & {0.8070 \textpm0.0012} & {176,599,584} \\
\hline
\multicolumn{3}{l}{\it ReQA NQ} \\
{Ours\textsubscript{sgpt}} & \textbf{0.6554 \textpm0.0007} & \textbf{4,609,808} \\
{-mining\textsubscript{sgpt}} & {0.6535 \textpm0.0005} & {149,379,552} \\
\hline
\end{tabular}
\caption{Differences in performance and training pair when unnecessary samples are omitted.}
\label{tab:table_6}
\end{table}

\subsection*{Low-resource Setting}
To emulate a low-resource scenario, we ran an experiment using only 10\% of the training data. The findings demonstrate that the method can work even in scenarios with limited data. This suggests that our method also works on a smaller training size, as shown in \autoref{tab:table_8}.

\begin{table}[H]
\centering
\begin{tabular}{r c c c}
\hline
\textbf{Model} & \textbf{Data size} & \textbf{MRR} & \textbf{R@1}\\
\hline
\multicolumn{4}{l}{\it ReQA SQuAD} \\
{Ours\textsubscript{use}} & {100\%} & \textbf{0.750} & \textbf{0.663}  \\
{Ours\textsubscript{use}} & {10\%} & {0.749} & {0.661}  \\
{Ours\textsubscript{mp}} & {100\%} & \textbf{0.748} & \textbf{0.658}  \\
{Ours\textsubscript{mp}} & {10\%} & {0.744} & {0.652}  \\
{Ours\textsubscript{sgpt}} & {100\%} & \textbf{0.815} & {0.736}  \\
{Ours\textsubscript{sgpt}} & {10\%} & \textbf{0.815} & \textbf{0.737}  \\
\hline
\multicolumn{4}{l}{\it ReQA NQ} \\
{Ours\textsubscript{use}} & {100\%} & {0.590} & {0.452}  \\
{Ours\textsubscript{use}} & {10\%} & \textbf{0.592} & \textbf{0.454}  \\
{Ours\textsubscript{mp}} & {100\%} & \textbf{0.642} & \textbf{0.457}  \\
{Ours\textsubscript{mp}} & {10\%} & {0.641} & {0.456}  \\
{Ours\textsubscript{sgpt}} & {100\%} & {0.655} & \textbf{0.529}  \\
{Ours\textsubscript{sgpt}} & {10\%} & \textbf{0.656} & \textbf{0.529}  \\
\hline
\end{tabular}
\caption{The performances of hybrid models (BM25-neural) on low-resource datasets.}
\label{tab:table_8}
\end{table}

\subsection*{Combining More Than Two Models}
Our framework also supports extensions beyond two models. We conducted experiments to combine three retrieval models. The results indicate that combining a model with a different specialization leads to a substantial performance boost, as demonstrated by the combination of SGPT-5.8B-msmarco, MPNET-QA, and BM25. This combination surpasses the current state-of-the-art in ReQA SQuAD, yielding an improvement from 0.801 to 0.823 on the MRR metric \cite{10082873}. The results are shown in \autoref{tab:table_9}.

\begin{table}[H]
\centering
\begin{tabular}{r c c c}
\hline
\textbf{Model} & \textbf{MRR} & \textbf{R@1} & \textbf{R@10}\\
\hline
\multicolumn{4}{l}{\it ReQA SQuAD} \\
{USE\textsubscript{mp+bm25}} & {0.784} & {0.697} & {0.926}  \\
{USE\textsubscript{sgpt+bm25}} & {0.807} & {0.732} & {0.930}  \\
{MPNET\textsubscript{use+bm25}} & {0.786} & {0.697} & {0.934}  \\
{MPNET\textsubscript{sgpt+bm25}} & {0.815} & {0.734} & {0.946}  \\
{SGPT\textsubscript{use+bm25}} & {0.819} & {0.741} & {0.948}  \\
{SGPT\textsubscript{mp+bm25}} & \textbf{0.823} & \textbf{0.742} & \textbf{0.957}  \\
\hline
\multicolumn{4}{l}{\it ReQA NQ} \\
{USE\textsubscript{mp+bm25}} & {0.700} & {0.555} & {0.920}  \\
{USE\textsubscript{sgpt+bm25}} & {0.663} & {0.537} & {0.867}  \\
{MPNET\textsubscript{use+bm25}} & {0.715} & {0.556} & {0.961}  \\
{MPNET\textsubscript{sgpt+bm25}} & \textbf{0.729} & \textbf{0.576} & \textbf{0.964}  \\
{SGPT\textsubscript{use+bm25}} & {0.670} & {0.545} & {0.876}  \\
{SGPT\textsubscript{mp+bm25}} & {0.714} & \textbf{0.576} & {0.922}  \\
\hline
\end{tabular}
\caption{Retrieval performances of trio models.}
\label{tab:table_9}
\end{table}

\subsection*{Performance on Modern QA Datasets}

We performed experiments using QA datasets from MTEB to assess the efficacy of our methods in modern retrieval tasks. We incorporated MPNET-QA and SGPT-125M-msmarco, utilizing the development set for training and the test set for evaluation. The results presented in \autoref{tab:table_10} show the effectiveness of our approach across datasets.

\begin{table}[H]
\centering
\begin{tabular}{r c c c}
\hline
\textbf{Dataset} & \textbf{MPNET} & \textbf{SGPT} & \textbf{Ours} \\
\hline
{DBPedia} & {0.656} & {0.264} & \textbf{0.669} \\
{FEVER} & {0.542} & {0.151} & \textbf{0.546} \\
{FiQA2018} & \textbf{0.560} & {0.120} & \textbf{0.560} \\
{HotpotQA} & {0.622} & {0.234} & \textbf{0.634} \\
{MSMARCO} & {0.352} & {0.425} & \textbf{0.663} \\
{NFCorpus} & \textbf{0.539} & {0.130} & \textbf{0.539} \\
{QuoraRetrieval} & {0.868} & {0.797} & \textbf{0.877} \\
\hline
\end{tabular}
\caption{Model performances on question answering datasets from MTEB, reported in MRR.}
\label{tab:table_10}
\end{table}

\subsection*{End-to-end Evaluation on RAG}
We conducted Retrieval-Augmented Generation (RAG) experiments to assess the end-to-end performance of our method using the mistral-7b-instruct-v0.2 \cite{jiang2023mistral} as a reader model. The outcomes were evaluated based on METEOR \cite{banerjee-lavie-2005-meteor}, BERT-F1 \cite{DBLP:conf/iclr/ZhangKWWA20}, and ROUGE-L scores. Our systems demonstrate performance improvements, as shown in \autoref{tab:table_12}.

\begin{table}[H]
\centering
\begin{tabular}{r c c c}
\hline
\textbf{Model} & \textbf{ME} & \textbf{BERT} & \textbf{RO-L} \\
\hline
\multicolumn{4}{l}{\it SQuAD} \\
{BM25} & {0.298} & {0.722} & {0.194}  \\
{MPNET} & {0.225} & {0.700} & {0.143}  \\
{SGPT} & {0.346} & {0.737} & {0.225}  \\
{MPNET + BM25} & {0.328} & {0.731} & {0.213}  \\
{SGPT + BM25} & {0.359} & {0.741} & {0.233}  \\
{SGPT + MPNET} & {0.348} & {0.738} & {0.227}  \\
{Trio} & \textbf{0.362} & \textbf{0.742} & \textbf{0.236}  \\
{Oracle Retriever} & {0.464} & {0.770} & {0.298}  \\
\hline
\multicolumn{4}{l}{\it NQ} \\
{BM25} & {0.251} & {0.697} & {0.155}  \\
{MPNET} & {0.286} & {0.706} & {0.173}  \\
{SGPT} & {0.325} & {0.719} & {0.200}  \\
{MPNET + BM25} & {0.289} & {0.707} & {0.175}  \\
{SGPT + BM25} & {0.325} & {0.719} & {0.201}  \\
{SGPT + MPNET} & {0.344} & \textbf{0.724} & {0.212}  \\
{Trio} & \textbf{0.345} & \textbf{0.724} & \textbf{0.213}  \\
{Oracle Retriever} & {0.362} & {0.742} & {0.236} \\
\hline
\end{tabular}
\caption{The performances of models in RAG tasks.}
\label{tab:table_12}
\end{table}

\subsection*{Multilingual Settings}
We conducted a non-English language experiment to assess the performance of our method when utilizing a multilingual model setup. We performed the experiment in the context of Thai language, utilizing the iapp-wiki-qa-squad datasets \cite{kobkrit_viriyayudhakorn_2021_4539916} which adhere to the SQuAD styles. Our modifications to support the Thai language include:
 
1) Changing USE-QA, which only supports English language, to Multilingual Universal Sentence Encoder Question Answering (mUSE-QA).

2) Substituting the NLTK \cite{bird-loper-2004-nltk} tokenizer with the PyThaiNLP \cite{phatthiyaphaibun-etal-2023-pythainlp} newmm tokenizer.

3) Using BM25 as the main retriever instead of the neural model due to a significant gap in performance on the R@10 metric.

The result shows that our method outperforms routing techniques. Notably, routing strategies only selected the ranking from the BM25 models, yielding a 0\% uplift in performance. The results are shown in \autoref{tab:table_11}.

\begin{table}[H]
\centering
\begin{tabular}{r c c c}
\hline
\textbf{Model} & \textbf{MRR} & \textbf{R@1} & \textbf{R@10}\\
\hline
\multicolumn{4}{l}{\it ReQA iAppTH} \\
{BM25} & {0.828} & {0.742} & {0.951}  \\
{mUSE\textsubscript{qa}} & {0.636} & {0.533} & {0.811}  \\
{Routing\textsubscript{muse}} & {0.828} & {0.742} & {0.951}  \\
{BM25\textsubscript{muse}} & \textbf{0.835} & \textbf{0.748} & \textbf{0.960}  \\
\hline
\end{tabular}
\caption{Model performances on multilingual settings.}
\label{tab:table_11}
\end{table}

\section{Conclusion} 

In this paper, we propose an effective method to combine term weighting and neural-based ReQA systems. Our approach involves training re-ranker neural networks using a modified pairwise re-ranking objective. Through empirical evaluations, the combined system exhibits a significant performance improvement over other combining strategies, particularly when confronted with significant performance disparities among the individual models. It is noteworthy that this technique is not restricted to a specific type or quantity of retrieval models. Our method focuses on improving the QA system by enhancing the retrieval which we identified as a bottleneck. Further investigation into improving the overall performance by directly optimizing the reader along with the retrieval model might improve the overall pipeline.

\section*{Limitations}

In our work, we have focused on implementing our methodology within the context of the retrieval question answering task. Nonetheless, it is reasonable to anticipate that the application of our techniques to a broader retrieval task would yield favorable results. Our methodology necessitated the selection of a main retriever. This is helpful in reducing the computing load of the supporting retriever, yet it can introduce a cap on the final performance. In the future, we may try to eliminate the need for main retrieval model selection. We also focus our experiment on techniques that do not require full-weight updates. Nevertheless, our approach can complement other full-weight update fine-tuning techniques to further enhance performance. Lastly, our model's computational cost scales with the number of re-ranking indexes fed through the re-ranker ($k$), as the process requires the prediction of $k\textsuperscript{2}$ pairs of attributes to determine the final ranking. This aspect may present challenges when deploying the model in situations with a tight compute budget.

\section*{Acknowledgements}
This work is under the joint research initiative between KASIKORN Business-Technology Group (KBTG) and the Faculty of Engineering, Chulalongkorn University. We would like to thank KASIKORNBANK PUBLIC COMPANY LIMITED (KBank) and KBTG for supporting the research and providing computational resources to conduct the experiments.

% Entries for the entire Anthology, followed by custom entries
% \newpage
\bibliography{anthology,custom}
\clearpage
\appendix

\label{sec:appendix}

\section{Effect of the Amount of Re-ranked Documents}

\begin{figure}[H]
\centering
\caption{Relationship between the number of re-ranked documents ($k$) and retrieval performance (MRR).}
\includegraphics[width=6.6cm]{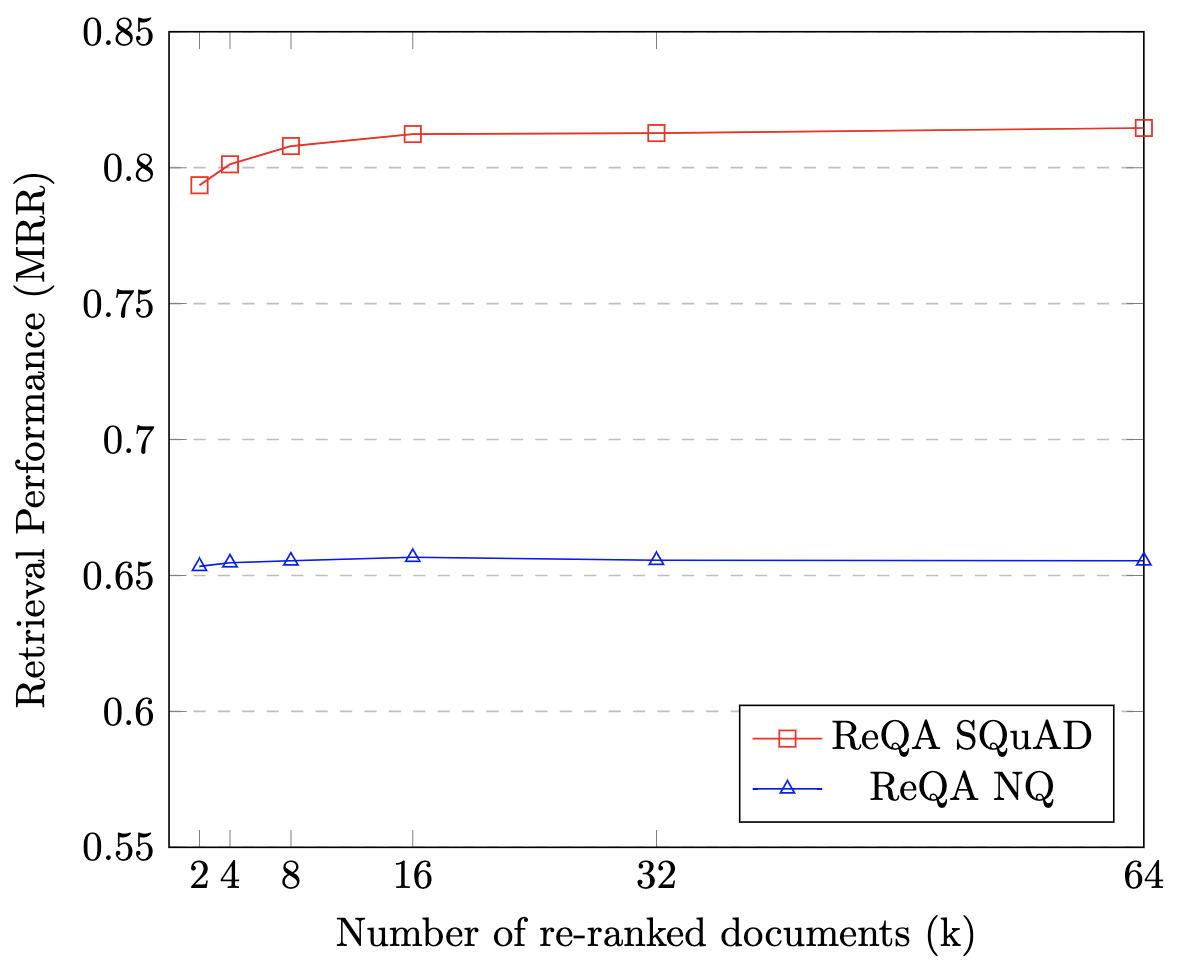}
\label{fig:figure_4}
\end{figure}

\begin{figure}[H]
\centering
\caption{Relationship between the number of re-ranked documents ($k$) and inference time (query/sec).}
\includegraphics[width=6.6cm]{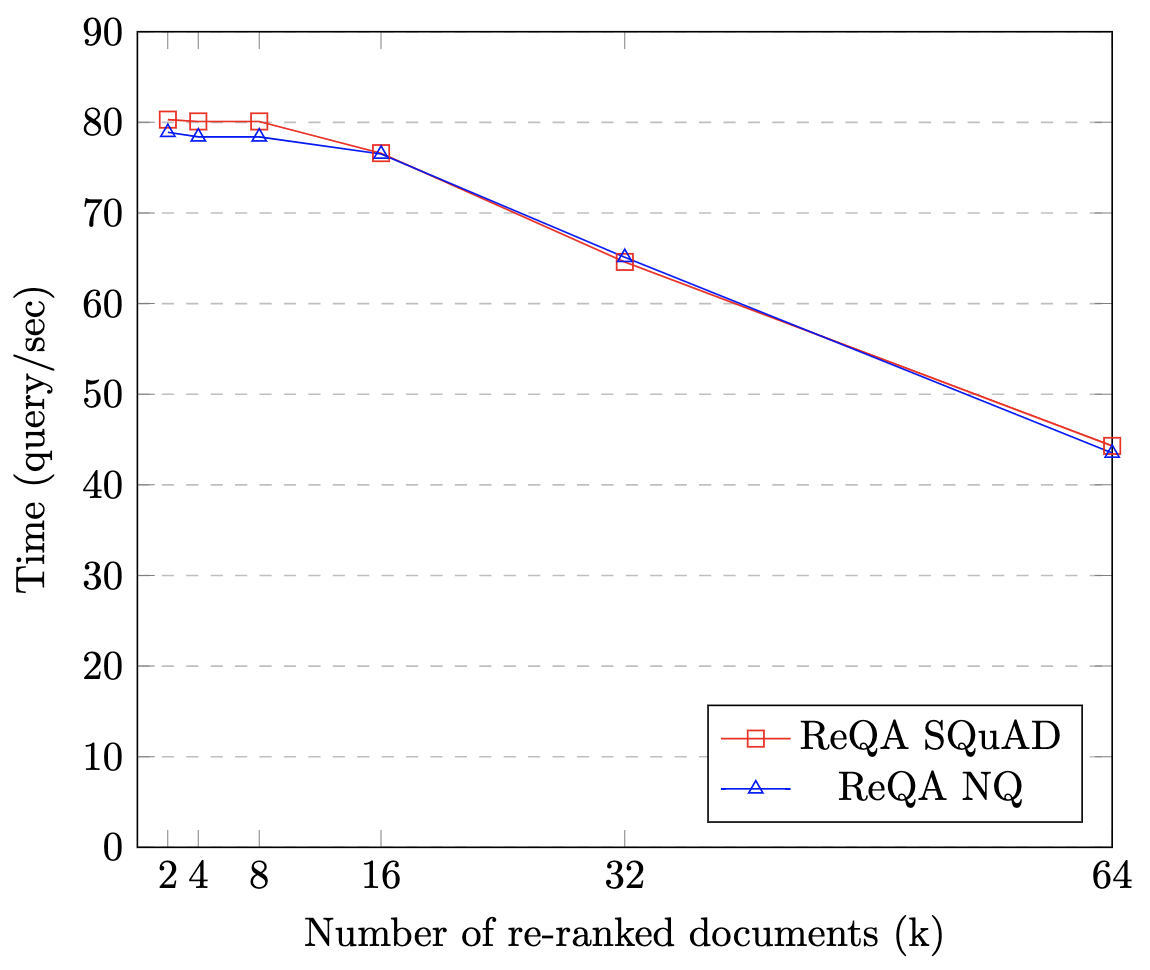}
\label{fig:figure_5}
\end{figure}

We conducted an experiment to study the effect of the number of re-ranked documents on the retrieval performance (\autoref{fig:figure_4}) and runtime (\autoref{fig:figure_5}). The processing was conducted on a standalone machine featuring Intel(R) Core(TM) i7-7700 CPU @ 3.60GHz, with no GPU acceleration.

To measure the latency, we ran our trio model with $k=16$. The re-ranker network adds a latency of 0.0133 seconds per query on average across datasets.

\newpage

\section{Correlation Between Retrieval and Overall QA Performance}

We assess the correlation between retrieval and overall QA performance. The findings reveal a strong correlation across metrics, suggesting that improvements in retrieval performance have a positive influence on overall QA performance.

\begin{table}[H]
\centering
\begin{tabular}{r c c c}
\hline
\textbf{Metric} & \textbf{MRR} & \textbf{R@1} \\
\hline
\multicolumn{3}{l}{\it SQuAD} \\
{ROUGE-L} & {0.8225} & {0.8392} \\
{EM} & {0.8182} & {0.8385} \\
\hline
\multicolumn{3}{l}{\it NQ} \\
{ROUGE-L} & {0.8806} & {0.8909} \\
{EM} & {0.8430} & {0.8751} \\
\hline
\end{tabular}
\caption{Correlation coefficient between retrieval and overall question answering performance.}
\label{tab:table_22}
\end{table}

\begin{table*}[t]
\centering
\begin{tabular}{r c c c c}
\hline
\textbf{Model} & \textbf{MRR} & \textbf{Recall@1} & \textbf{Recall@5} & \textbf{Recall@10}\\
\hline
\multicolumn{1}{l}{\it ReQA SQuAD} & {} & {} & {} & {}  \\
{Ours\textsubscript{use}} & {0.7503 \textpm0.0002} & {0.6629 \textpm0.0002} & {0.8598 \textpm0.0006} & {0.9015 \textpm0.0004} \\
{Ours\textsubscript{mpnet}} & {0.7476 \textpm0.0004} & {0.6576 \textpm0.0008} & {0.8561 \textpm0.0004} & {0.9090 \textpm0.0003} \\
{Ours\textsubscript{sgpt}} & {0.8146 \textpm0.0004} & {0.7364 \textpm0.0005} & {0.9131 \textpm0.0006} & {0.9451 \textpm0.0005} \\
\hline
\multicolumn{1}{l}{\it ReQA NQ} & {} & {} & {} & {}  \\
{Ours\textsubscript{use}} & {0.5902 \textpm0.0012} & {0.4518 \textpm0.0022} & {0.7646 \textpm0.0005} & {0.8319 \textpm0.0008} \\
{Ours\textsubscript{mpnet}} & {0.6422 \textpm0.0014} & {0.4568 \textpm0.0026} & {0.9077 \textpm0.0008} & {0.9510 \textpm0.0006} \\
{Ours\textsubscript{sgpt}} & {0.6554 \textpm0.0007} & {0.5292 \textpm0.0011} & {0.8160 \textpm0.0006} & {0.8621 \textpm0.0019} \\
\hline
\end{tabular}
\caption{Summary statistics for experimental results on ReQA SQuAD and ReQA NQ.}
\end{table*}

\begin{table*}[t]
\centering
\begin{tabular}{|p{1.5in}|p{2.0in}|p{2.0in}|}
\hline
\textbf{Query} & \textbf{Routing} & \textbf{Ours}\\
\hline
{What two artists came out with Coldplay during the half-time show?} & {On December 3, the league confirmed that the show would be headlined by the British rock group Coldplay.} & {The Super Bowl 50 halftime show was headlined by the British rock group Coldplay with special guest performers Beyoncé and Bruno Mars, who headlined the Super Bowl XLVII and Super Bowl XLVIII halftime shows, respectively.} \\
\hline
{What type of relationships do enthusiastic teachers cause?} & {Students who had enthusiastic teachers tend to rate them higher than teachers who didn't show much enthusiasm for the course materials.} & {Enthusiastic teachers are particularly good at creating beneficial relations with their students.} \\
\hline
{When did Luther return to Wittenberg?} & {When the town council asked Luther to return, he decided it was his duty to act.} & {Luther secretly returned to Wittenberg on 6 March 1522.} \\
\hline
{What is the mortality rate of pneumonic plague?} & {Septicemic plague is the least common of the three forms, with a mortality rate near 100\%.} & {Pneumonic plague has a mortality rate of 90 to 95 percent.}\\
\hline
{What was the population Jacksonville city as of 2010?} & {Jacksonville has Florida's largest Filipino American community, with 25,033 in the metropolitan area as of the 2010 Census.} & {Jacksonville is the principal city in the Jacksonville metropolitan area, with a population of 1,345,596 in 2010.} \\
\hline
{What was the result of the 1967 referendum?} & {However, as a result of the referendum in France and the referendum in the Netherlands, the 2004 Treaty establishing a Constitution for Europe never came into force.} & {When a consolidation referendum was held in 1967, voters approved the plan.} \\
\hline
\end{tabular}
\caption{Comparative analysis between Routing and our model reveals distinct characteristics. The outcomes obtained from the Routing technique demonstrate the limitations inherent in the BM25 model, as it heavily relies on the presence of overlapping evidence while inadequately addressing the critical aspect of question-answer alignment. Conversely, our model exhibits a robust approach by incorporating ranking information from the neural model.}
\end{table*}

\end{document}